\documentclass[conference]{IEEEtran}
\IEEEoverridecommandlockouts
\usepackage{cite}
\usepackage{amsmath,amssymb,amsfonts}
\usepackage{algorithmic}
\usepackage{graphicx}
\usepackage{textcomp}
\usepackage{xcolor}
\usepackage{url}
\def\BibTeX{{\rm B\kern-.05em{\sc i\kern-.025em b}\kern-.08em
    T\kern-.1667em\lower.7ex\hbox{E}\kern-.125emX}}
\begin{document}

\title{Subject-Level Unknown-Identity Identification from Leap Motion Controller 2 Hand Landmarks \\
% {\footnotesize \textsuperscript{*}Note: Sub-titles are not captured in Xplore and should not be used}
}

\author{\IEEEauthorblockN{Bahar Moharrer}
\IEEEauthorblockA{\textit{Department of Computer Science} \\
\textit{Sapienza University of Rome}\\
Rome, Italy \\
moharrer.2058349@studenti.uniroma1.it}
\and
\IEEEauthorblockN{Susanna Cifani}
\IEEEauthorblockA{\textit{Department of Computer Science} \\
\textit{Sapienza University of Rome}\\
Rome, Italy \\
cifani@di.uniroma1.it}
\and
\IEEEauthorblockN{Marco Raoul Marini}
\IEEEauthorblockA{\textit{Department of Computer Science} \\
\textit{Sapienza University of Rome}\\
Rome, Italy \\
marini@di.uniroma1.it}
\and
\IEEEauthorblockN{Luigi Cinque}
\IEEEauthorblockA{\textit{Department of Computer Science} \\
\textit{Sapienza University of Rome}\\
Rome, Italy \\
cinque@di.uniroma1.it}
\and
\IEEEauthorblockN{Maria De Marsico}
\IEEEauthorblockA{\textit{Department of Computer Science} \\
\textit{Sapienza University of Rome}\\
Rome, Italy \\
demarsico@di.uniroma1.it}
}
% \author{
% \IEEEauthorblockN{Author One, Author Two, Author Three, Author Four, and Author Five}
% \IEEEauthorblockA{\textit{Department / Institution, City, Country}\\
% email1, email2, email3, email4, email5}
% }

% \author{\IEEEauthorblockN{Anonym for Double Blind}}

\maketitle

\begin{abstract}
This work studies subject recognition from Leap Motion Controller 2 (LMC2) hand landmark data under a subject-level unknown-identity identification protocol on the Multi View Leap2 Hand Pose (ML2HP) dataset. Using only the landmark modality, we retain the original geometric representation and enrich it with fingertip-to-palm distances and palm-normalized inter-finger angular descriptors. Evaluation is performed under a Leave-One-Subject-Out (LOSO) protocol in which, for each outer fold, one subject is excluded from the enrolled set and treated as unknown at test time. To avoid tuning on the true outer unknown subject, the unknown-rejection threshold is selected in an inner validation step by temporarily withholding one enrolled subject from the inner gallery and using it only for threshold estimation.

We compare a tree ensemble baseline with two neural alternatives: a learned embedding baseline based on centroid matching and cosine-similarity-based rejection, and an MLP+OpenMax model, which represents a more established open-set recognition approach. Under this evaluation setup, Extra Trees remains the strongest overall method, indicating that the main challenge on this benchmark is not enrolled-subject discrimination alone, but robust score separation between known and unknown probes. The results support the feasibility of compact, interpretable landmark-based descriptors for contactless hand-based unknown-subject rejection and identification on a small-cohort dataset.

% This document is a model and instructions for \LaTeX.
% This and the IEEEtran.cls file define the components of your paper [title, text, heads, etc.]. *CRITICAL: Do Not Use Symbols, Special Characters, Footnotes, or Math in Paper Title or Abstract.
\end{abstract}

\begin{IEEEkeywords}
Hand biometrics, open-set recognition, subject identification, Leap Motion Controller 2, hand landmarks, LOSO evaluation
\end{IEEEkeywords}

\section{Introduction}

Hand analysis is commonly studied for pose estimation, gesture recognition, or sign language understanding, where the goal is to recognize what the hand is doing. Biometric subject recognition, in contrast, asks who produced the sample, shifting the emphasis from pose invariance to anatomy-related geometric differences. Landmark-based sensing is attractive for this task because it provides a compact, structured, and interpretable description of hand geometry. Leap Motion Controller 2 (LMC2)\footnote{\url{https://docs.ultraleap.com/}} sensors directly capture palm pose, joint locations, finger widths, and related geometric descriptors without requiring wearable devices. The Multi-view Leap2 Hand Pose (ML2HP) dataset \cite{ML2HP_Dataset,DLMC2_DS_paper}, originally introduced for hand-pose recognition, is therefore suitable for reformulation as a landmark-based biometric benchmark.

In this study, we adopt a subject-level unknown-identity identification formulation under a Leave-One-Subject-Out (LOSO) protocol. In each outer fold, one subject is excluded from the enrolled set and treated as unknown at test time, while the remaining subjects form the enrolled identities. This setting evaluates both enrolled-subject recognition and rejection of an unseen, unenrolled identity on a small-cohort landmark benchmark.
The contributions are threefold. First, we reformulate ML2HP as a landmark-only subject-identification benchmark with explicit unknown-subject rejection. Second, we evaluate compact geometric descriptors under a nested threshold-selection procedure that avoids tuning on the true held-out unknown subject. Third, we compare a non-linear tree ensemble, an embedding MLP with centroid-based rejection, and an MLP+OpenMax model under the same protocol, isolating the effect of the scoring model rather than differences in modality or input representation.

\section{Related Work}
\label{sec:relatedwork}

Open-set recognition requires a model to recognize known classes while rejecting samples from classes absent during training. Surveys by Geng et al.~\cite{Geng2018RecentAI} and Sun and Dong~\cite{sun2023surveyopensetimgrec} distinguish this setting from standard closed-set classification and emphasize the need to evaluate both known-class recognition and unknown-class rejection. Among deep open-set methods, OpenMax remains a common reference baseline: Bendale and Boult~\cite{bendale2016openmax} proposed recalibrating activation-based class scores to estimate an explicit unknown probability. More recently, Vaze et al.~\cite{vaze2022openset} showed that strong closed-set classifiers combined with simple scoring rules can be competitive for open-set recognition, which is relevant here because our best-performing method is also the strongest closed-set classifier. Score calibration is also important for threshold-based rejection, as neural confidence scores are often poorly calibrated~\cite{guo2017calibration}.

Open-set evaluation is especially important in biometrics, where a probe may correspond to either an enrolled or unenrolled identity. Su et al.~\cite{su2024openset} argued that strong closed-set biometric models do not necessarily yield strong open-set performance, because conventional objectives are not explicitly aligned with unknown-identity rejection. In the present work, openness is defined at the subject level: in each LOSO fold, the unknown identity is not only absent from the enrolled gallery but also excluded from model fitting. The resulting protocol should therefore be interpreted as held-out-subject unknown-identity identification rather than as a large-scale biometric deployment benchmark.

Leap Motion hand data has previously been used for authentication and person recognition. Chan et al.~\cite{Chan2015LeapMotion}, Maruyama et al.~\cite{Maruyama2017}, and Shin et al.~\cite{Shin2022DTW,Shin2024Dynamic} showed that hand geometry and dynamic gesture information can support user authentication or identification, but their settings differ from landmark-only open-set subject identification under LOSO evaluation. Landmark-based hand analysis has also been studied for gesture and pose recognition; Gil-Mart\'{i}n et al.~\cite{icaart2025MediaPipe} showed that normalized landmarks and anthropometric descriptors can be effective, while Esteban-Romero et al.~\cite{icaart2025TowardsMV} used the ML2HP dataset under LOSO evaluation for multi-view hand pose recognition using both image embeddings and Leap 2 landmarks. In contrast, this work reformulates ML2HP as a landmark-only subject-identification benchmark with explicit unknown-identity rejection.

\section{Methodology}
\label{sec:methodology}
\begin{figure*}
    \centering
    \includegraphics[width=0.8\linewidth]{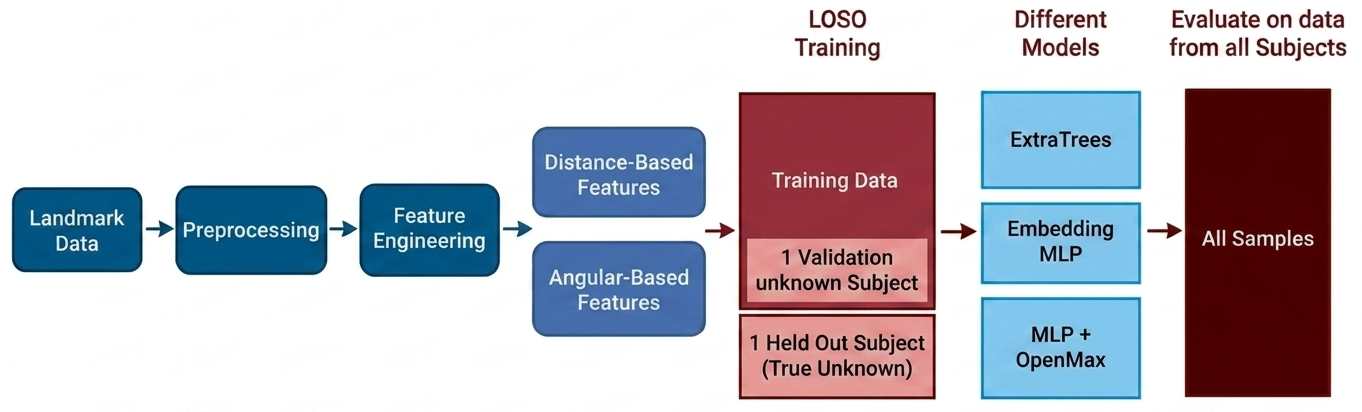}
    \caption{Overall architecture of the pipeline: Preprocessing, Feature Engineering, Classification.}
    \label{fig:Architecture}
\end{figure*}

Our proposed pipeline for LOSO open-set subject identification follows a structured three-stage approach: data preprocessing, geometric feature engineering, and classification under an open-set protocol. In each fold, one subject is excluded from training and treated as an unknown identity at test time, while the remaining subjects form the enrolled set. Figure \ref{fig:Architecture} provides an overview of the pipeline, while the LOSO open-set protocol is detailed in Section \ref{LOSO Protocol}.

\subsection{Dataset}

Experiments are conducted on the ML2HP dataset, which contains 21 subjects performing 17 static hand poses captured using two perpendicular LMC2 devices. The full release contains 714,000 synchronized records, each with a 247-dimensional landmark representation \cite{DLMC2_DS_paper}. In this work, only the landmark modality is used.

The landmarks include palm position and orientation, joint-level coordinates, finger widths, and hand-level descriptors, making the dataset suitable for studying intrinsic hand geometry without image processing. Following the original dataset structure, Horizontal and Vertical device records are processed as device-specific landmark samples rather than as a pre-concatenated multi-view vector.

Because the dataset contains only 21 identities, the study should be interpreted as a pilot benchmark for landmark-only hand biometrics.

\subsection{Preprocessing}
We first perform landmark data cleaning to ensure consistency. Missing values and duplicate entries are removed. Metadata columns that do not directly describe hand geometry, such as pose labels, frame identifiers, device descriptors, and other administrative fields, are excluded from the input feature matrix. The subject identifier is retained only as the target identity label and is never used as an input feature. After cleaning, the data are converted into a numerical feature matrix containing only informative landmark variables.

\subsection{Feature Engineering}

Raw LMC2 landmarks provide detailed geometric information, but additional descriptors can better emphasize subject-specific hand structure. We therefore augment the original landmark representation with two complementary feature groups: fingertip-to-palm distances and palm-normalized angular descriptors. The distance features capture global hand proportions, while the angular features describe relative finger configurations in a local hand-centered coordinate system.

\subsubsection{Distance-based Features}

For each sample, the Euclidean distance between the palm center and each fingertip is computed for the thumb, index, middle, ring, and pinky fingers:
\begin{equation}
d_f =
\sqrt{(x_f-x_p)^2 + (y_f-y_p)^2 + (z_f-z_p)^2},
\label{eq:distance}
\end{equation}
where $(x_f,y_f,z_f)$ and $(x_p,y_p,z_p)$ denote the 3D coordinates of fingertip $f$ and the palm center, respectively. This yields five scalar descriptors that summarize fingertip positions relative to the palm.

\subsubsection{Palm-normalized Angle Features}

Angle-based features are computed from finger direction vectors expressed in a local palm coordinate system. For each finger, the direction vector is defined from its base joint to its fingertip. Let $\mathbf{d}$ be the palm direction vector, $\mathbf{n}$ the palm normal vector, and $\mathbf{r}=\mathbf{d}\times\mathbf{n}$ the orthogonal right-hand vector. These three unit vectors define the palm-centered frame, and each finger vector is projected into this frame before angle computation. This normalization makes the angular descriptors less sensitive to global hand translation and rotation, so the resulting features better reflect intrinsic hand geometry.

For each pair of normalized finger vectors $\mathbf{v}'_{f_i}$ and $\mathbf{v}'_{f_j}$, the pairwise angle is computed as
\begin{equation}
\theta_{ij} =
\arccos\left(
\frac{\mathbf{v}'_{f_i}\cdot \mathbf{v}'_{f_j}}
{\|\mathbf{v}'_{f_i}\|\,\|\mathbf{v}'_{f_j}\|}
\right),
\label{eq:pairwise_angle}
\end{equation}
producing ten inter-finger angle features per sample.

To capture higher-order finger geometry, we also compute triplet angles from three fingertip points $A$, $B$, and $C$, with the angle measured at $B$:
\begin{equation}
\theta_{ABC} =
\arccos\left(
\frac{(A-B)\cdot(C-B)}
{\|A-B\|\,\|C-B\|}
\right).
\label{eq:triplet_angle}
\end{equation}
These descriptors capture local finger-group configurations beyond pairwise spread. The final feature vector concatenates the raw landmark attributes, five fingertip-to-palm distances, and the pairwise and triplet angular descriptors.

\subsection{Subject-level LOSO protocol with inner threshold selection}
\label{LOSO Protocol}

The evaluation follows a subject-level LOSO protocol with an inner validation step for threshold selection. In each outer fold, all samples from one subject are excluded from the enrolled set and treated as the true unknown test identity. The remaining subjects constitute the outer enrolled identities. A classifier is then trained on the outer enrolled subjects and evaluated on the complete dataset, so that samples from enrolled subjects are treated as known, and those from the held-out subject are treated as unknown.

To select the unknown-rejection threshold without tuning on the true outer unknown subject, we use an inner validation procedure within the outer enrolled set. Specifically, one enrolled subject is temporarily withheld from the inner gallery and used solely for validation of unknown identities, while the remaining enrolled subjects are used for model fitting. Validation scores are then computed for two groups: known validation samples from the inner enrolled subjects, and unknown validation samples from the temporarily withheld enrolled subjects. The rejection threshold is selected on this inner validation split using the equal-error-rate criterion and is then applied unchanged in the corresponding outer fold.

This temporarily withheld validation subject is known at the dataset level, but it is excluded from the inner gallery and treated as unknown only for threshold selection. The purpose of this step is not to simulate a large-scale biometric deployment, but to avoid tuning the threshold directly on the true held-out outer subject.

For Extra Trees, the unknown score is defined as
\begin{equation}
s_{\text{unk}} = 1 - p_{\max},
\label{eq:et_unknown_score}
\end{equation}
where $p_{\max}$ is the maximum predicted posterior over the enrolled identities.

For the embedding baseline, the model outputs an $\ell_2$-normalized feature vector $\mathbf{z}$, and one centroid $\mathbf{c}_k$ is computed for each enrolled identity $k$ in embedding space. Cosine similarity is then measured between a probe embedding and each enrolled centroid, and the unknown score is defined as
\begin{equation}
s_{\text{unk}} = 1 - \max_k \left( \frac{\mathbf{z} \cdot \mathbf{c}_k}{\|\mathbf{z}\| \, \|\mathbf{c}_k\|} \right),
\label{eq:embed_unknown_score}
\end{equation}
where the maximum is taken over all enrolled identities.

For the OpenMax model, an MLP classifier is first trained on the enrolled identities, class-wise mean activation vectors and Weibull tail models are then estimated from the training activations, and the unknown score is taken as the OpenMax unknown probability after score recalibration:
\begin{equation}
s_{\text{unk}} = p_{\text{OpenMax}}(\text{unknown} \mid \mathbf{x}),
\label{eq:openmax_unknown_score}
\end{equation}
where $\mathbf{x}$ denotes the input landmark feature vector. In all cases, higher unknown-score values indicate stronger evidence for rejecting a probe as unknown.

This protocol produces three complementary result views. First, closed-set accuracy is computed only on enrolled identities. Second, open-set accuracy, macro F1-score, and unknown recall are computed over the enrolled identities plus the held-out unknown class. Third, a binary rejection analysis treats held-out-subject samples as positives and enrolled-subject samples as negatives; therefore, AUC, EER, FAR, and FRR measure known-versus-unknown score separation rather than full biometric verification performance.

\subsection{Classification Models}

We evaluate three classifiers under the same landmark-only open-set protocol. Extra Trees uses 200 estimators, min\_samples\_leaf=2, max\_features=sqrt, and random\_state=42, serving as a strong non-linear ensemble baseline.

The two neural baselines use standardized features and the same compact MLP backbone with hidden layers of 256 and 128 units, dropout 0.1, and a 64-dimensional embedding/activation layer. Training uses Adam with learning rate $10^{-3}$, weight decay $10^{-4}$, batch size 256, and 30 epochs. For the embedding baseline, class centroids are computed from L2-normalized embeddings, and probes are assigned by maximum cosine similarity. For MLP+OpenMax, class-wise mean activation vectors and Weibull tail models are fitted from training activations, and OpenMax recalibration provides the unknown probability used for rejection.

All models are evaluated on the same landmark-only feature set, so performance differences reflect the classifiers’ modeling capabilities rather than differences in the input modality.

\section{Results and Discussion}

\subsection{Main Results}

Table \ref{tab:main_results} summarizes the main recognition results for the three evaluated models and Table \ref{tab:binary_rejection} reports the corresponding binary unknown subject rejection metrics. All values are reported as mean $\pm$ standard deviation across LOSO folds.

\begin{table*}[t]
\centering
\caption{Main comparison of the evaluated models across LOSO folds. Values are reported as mean $\pm$ standard deviation.}
\setlength{\tabcolsep}{4pt}
\begin{tabular}{lcccc}
\hline
\textbf{Model} &
\begin{tabular}{@{}c@{}}\textbf{Closed-set}\\ \textbf{Accuracy}\end{tabular} &
\begin{tabular}{@{}c@{}}\textbf{Open-set}\\ \textbf{Accuracy}\end{tabular} &
\begin{tabular}{@{}c@{}}\textbf{Open-set}\\ \textbf{Macro F1}\end{tabular} &
\begin{tabular}{@{}c@{}}\textbf{Unknown Recall}\\ \textbf{(Rejection Rate)}\end{tabular} \\
\hline
Extra Trees & $99.49 \pm 0.06$\% & $91.29 \pm 1.50$\% & $92.90 \pm 1.02$\% & $85.13 \pm 7.71$\% \\
Embedding MLP + centroids & $94.78 \pm 0.25$\% & $74.23 \pm 1.66$\% & $81.04 \pm 1.21$\% & $73.42 \pm 6.70$\% \\
MLP + OpenMax & $95.10 \pm 0.28$\% & $74.32 \pm 1.62$\% & $81.10 \pm 1.18$\% & $73.39 \pm 7.80$\% \\
\hline
\end{tabular}

\label{tab:main_results}
\end{table*}

\begin{table*}[t]
\centering
\caption{Binary unknown subject rejection performance of the evaluated models across LOSO folds. (mean $\pm$ standard deviation).}
\setlength{\tabcolsep}{5pt}
\begin{tabular}{lcccc}
\hline
\textbf{Model} & \textbf{AUC} & \textbf{EER} & \textbf{FAR} & \textbf{FRR} \\
\hline
Extra Trees & $95.71 \pm 1.93$\% & $10.77 \pm 2.84$\% & $8.32 \pm 1.70$\% & $14.87 \pm 7.71$\% \\
Embedding MLP + centroids & $80.96 \pm 4.04$\% & $25.88 \pm 3.27$\% & $25.23 \pm 1.86$\% & $26.58 \pm 6.70$\% \\
MLP + OpenMax & $80.70 \pm 4.58$\% & $25.95 \pm 3.74$\% & $25.38 \pm 1.80$\% & $26.61 \pm 7.80$\% \\
\hline
\end{tabular}

\label{tab:binary_rejection}
\end{table*}

Under this evaluation setup, Extra Trees remains the strongest overall method. It achieves $99.49 \pm 0.06$\% closed-set accuracy, $91.29 \pm 1.50$\% overall open-set accuracy, $92.90 \pm 1.02$\% open-set macro F1, and $85.13 \pm 7.71$\% unknown recall. The gap relative to the neural models is substantial, indicating that the proposed landmark representation is especially effective when paired with a strong non-linear ensemble.

The two neural approaches perform similarly in overall open-set recognition, but with different trade-offs. The embedding MLP reaches $94.78 \pm 0.25$\% closed-set accuracy, $74.23 \pm 1.66$\% open-set accuracy, $81.04 \pm 1.21$\% open-set macro F1, and $73.42 \pm 6.70$\% unknown recall. The MLP+OpenMax model attains slightly higher closed-set accuracy, at $95.10 \pm 0.28$\%, and a nearly identical overall open-set accuracy of $74.32 \pm 1.62$\%, with $81.10 \pm 1.18$\% open-set macro F1 and $73.39 \pm 7.80$\% unknown recall. Thus, although OpenMax introduces an explicit open-set recognition mechanism, it does not close the gap to Extra Trees on this benchmark.

The rejection metrics in Table \ref{tab:binary_rejection} reinforce this interpretation. Extra Trees yields the strongest known-versus-unknown separation, achieving $95.71 \pm 1.93$\% AUC and $10.77 \pm 2.84$\% EER, with much lower FAR and FRR than either neural baseline. By contrast, the embedding MLP and OpenMax are clearly weaker in binary rejection, with AUC values of $80.96 \pm 4.04$\% and $80.70 \pm 4.58$\%, respectively, and EER values of $25.88 \pm 3.27$\% and $25.95 \pm 3.74$\%. This indicates that the main challenge in the present setting is not only enrolled-subject discrimination but also the production of stable, well-separated rejection scores for unseen identities.

Extra Trees was also the fastest model in our implementation, requiring 132 min compared with 319 min for the embedding MLP and 394 min for MLP+OpenMax.

\subsection{Statistical significance analysis}

To assess whether the observed performance differences were consistent across LOSO folds, paired Wilcoxon signed-rank tests were applied to the fold-wise results for overall open-set accuracy, open-set macro F1-score, AUC, and EER. Since all methods were evaluated on the same 21 held-out-subject folds, each comparison was treated as paired.

The results confirm that Extra Trees is consistently stronger than both neural baselines. Extra Trees significantly outperformed the embedding MLP on all four metrics ($p = 9.54 \times 10^{-7}$ in each case). It also significantly outperformed MLP+OpenMax on open-set accuracy and EER ($p = 9.54 \times 10^{-7}$), as well as on macro F1-score and AUC ($p = 5.95 \times 10^{-5}$). In contrast, no statistically significant difference was observed between the embedding MLP and MLP+OpenMax for open-set accuracy ($p = 0.6333$), macro F1-score ($p = 0.6827$), AUC ($p = 0.7335$), or EER ($p = 0.9729$). These results indicate that the advantage of Extra Trees is consistent across held-out subjects, whereas the two neural open-set baselines are statistically comparable in this setting.

%Overall, these results show that the stronger average performance of Extra Trees is not due to a small number of favorable folds, but is consistently observed across the LOSO evaluation.

\subsection{Unknown-score Analysis}

\begin{figure*}
    \centering
    \includegraphics[width=1\linewidth]{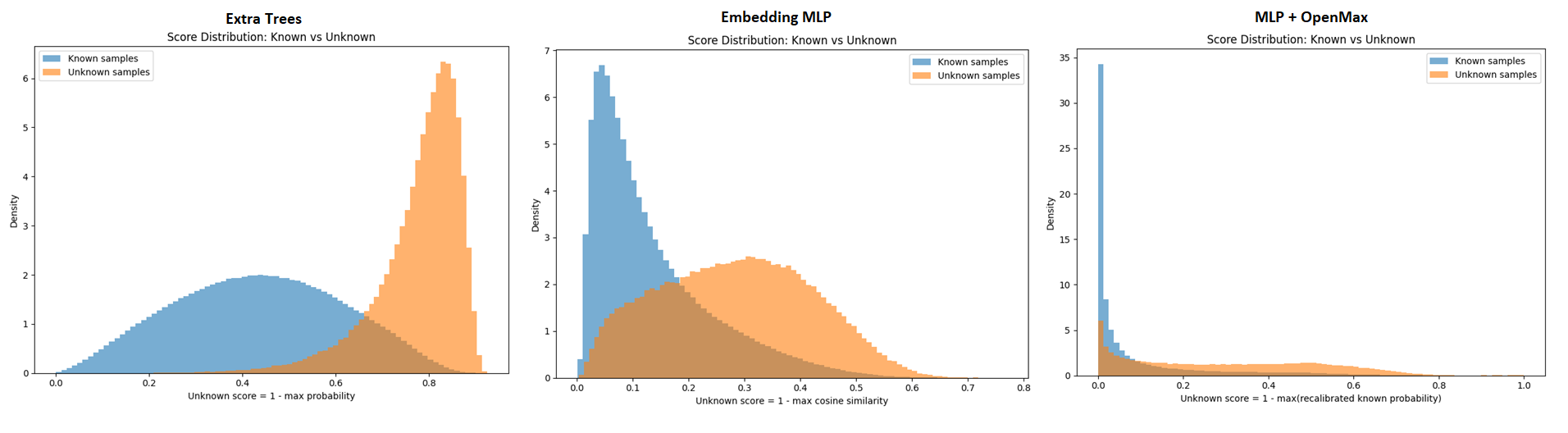}
    \caption{Unknown-score distributions for known and unknown probes under the LOSO open-set protocol: Extra Trees, using $1-p_{\max}$; Embedding MLP + centroids, using $1-\max(\mathrm{cosine\ similarity})$; and MLP + OpenMax, using the recalibrated unknown probability.}
    \label{fig:score_distributions}
\end{figure*}

Figure \ref{fig:score_distributions} compares the unknown score distributions of known and unknown probes for the three evaluated models: (a) Extra Trees, (b) Embedding MLP + centroids, and (c) MLP + OpenMax. Extra Trees shows the clearest separation, with known probes concentrated at lower unknown score values and unseen probes shifted toward higher values. The overlap region is relatively limited, consistent with its higher AUC and lower EER.

By contrast, the embedding MLP and MLP+OpenMax show greater overlap between known and unknown samples. In the embedding baseline, cosine similarity-based rejection provides weaker separation between enrolled and unseen identities. OpenMax produces a more concentrated distribution for known probes near low unknown-score values, but the unknown distribution still overlaps substantially with the known distribution. Overall, these score patterns support the quantitative results in Tables 1 and 2 and indicate that the advantage of Extra Trees lies not only in closed-set recognition but also in stronger known-versus-unknown score separation.

\subsection{Comparison with Related Work}

The following comparisons should be interpreted as contextual rather than directly equivalent, since prior studies differ in task definition, sensing modality, and evaluation protocol. A particularly relevant reference is the recent ML2HP baseline by Esteban-Romero et al. \cite{icaart2025TowardsMV}, as it uses the same dataset and a LOSO evaluation setting. However, that work addresses multi-view hand pose recognition rather than biometric subject identification, and combines grayscale imagery with landmark information. In contrast, the present study focuses specifically on landmark-only open-set subject identification, in which the system must both recognize enrolled users and reject identities not observed during training. Accordingly, the comparison is useful for gauging the benchmark's difficulty, but not as a direct numerical reference.
Earlier Leap Motion biometric studies by Chan et al. \cite{Chan2015LeapMotion}, and Shin et al. \cite{Shin2024Dynamic} likewise confirm the discriminative value of hand geometry and gesture information, but they address authentication or closed-set person identification across different datasets and protocols. In this context, the most relevant direct external comparison is the inclusion of OpenMax within the present study, since it provides an established open-set baseline evaluated under the same experimental setting.

\subsection{Ablation Study}

To assess the contribution of the engineered feature groups, we conducted an ablation study using four input configurations: \emph{raw landmarks}, \emph{raw landmarks + angle features}, \emph{raw landmarks + distance features}, and the \emph{full feature set}. Table \ref{tab:ablation_features} reports the results for all three models in terms of mean $\pm$ standard deviation across LOSO folds for closed-set accuracy and overall open-set accuracy.

\begin{table*}[t]
\centering
\small
\caption{Ablation study across feature configurations for the three models. (mean $\pm$ standard deviation across LOSO folds).}

\setlength{\tabcolsep}{5pt}
\begin{tabular}{llcc}
\hline
\textbf{Model} & \textbf{Feature Configuration} & \textbf{Closed-set Accuracy} & \textbf{Open-set Accuracy} \\
\hline
Extra Trees & Raw landmarks & $99.31 \pm 0.05$\% & $88.11 \pm 3.00$\% \\
 & Raw + angle features & $99.45 \pm 0.05$\% & $88.94 \pm 2.75$\% \\
 & Raw + distance features & $99.33 \pm 0.06$\% & $88.27 \pm 3.00$\% \\
 & Full feature set & $99.49 \pm 0.06$\% & $91.29 \pm 1.50$\% \\
\hline
Embedding MLP & Raw landmarks & $84.12 \pm 0.44$\% & $62.65 \pm 1.30$\% \\
 & Raw + angle features & $88.23 \pm 0.34$\% & $67.23 \pm 1.51$\% \\
 & Raw + distance features & $86.14 \pm 0.61$\% & $66.45 \pm 1.64$\% \\
 & Full feature set & $94.78 \pm 0.25$\% & $74.23 \pm 1.66$\% \\
\hline
MLP + OpenMax & Raw landmarks & $82.41 \pm 0.50$\% & $57.47 \pm 2.68$\% \\
 & Raw + angle features & $87.22 \pm 0.67$\% & $64.47 \pm 1.98$\% \\
 & Raw + distance features & $86.13 \pm 0.58$\% & $62.36 \pm 2.11$\% \\
 & Full feature set & $95.10 \pm 0.28$\% & $74.32 \pm 1.62$\% \\
\hline
\end{tabular}

\label{tab:ablation_features}
\end{table*}

For Extra Trees, the angle features provide the greatest individual improvement over raw landmarks, increasing open-set accuracy from $88.11 \pm 3.00$\% to $88.94 \pm 2.75$\%, whereas distance features alone produce only a marginal gain to $88.27 \pm 3.00$\%. The full Extra Trees model remains the strongest overall, reaching $99.49 \pm 0.06$\% closed-set accuracy and $91.29 \pm 1.50$\% open-set accuracy. This indicates that both engineered feature groups are useful, but the angle descriptors contribute more strongly when used in isolation.

A similar pattern is observed for the neural models. For the embedding MLP, using only raw landmarks yields $84.12 \pm 0.44$\% closed-set accuracy and $62.65 \pm 1.30$\% open-set accuracy. Adding either angle or distance features improves performance, but the full feature set produces the best result, reaching $94.78 \pm 0.25$\% closed-set accuracy and $74.23 \pm 1.66$\% open-set accuracy. The MLP+OpenMax model follows the same trend: the raw landmark configuration attains $82.41 \pm 0.50$\% closed-set accuracy and $57.47 \pm 2.68$\% open-set accuracy, while the full feature set increases performance to $95.10 \pm 0.28$\% and $74.32 \pm 1.62$\%, respectively.

Overall, the ablation study supports two main conclusions. First, the handcrafted geometric descriptors consistently improve performance beyond raw landmark coordinates alone, with the angle-based features providing the strongest single contribution. Second, although both neural models benefit markedly from the engineered descriptors, Extra Trees remains the most robust across all feature configurations. This suggests that the proposed geometric representation is effective for all three models, but particularly well matched to the decision boundaries learned by the tree ensemble.

\section{Conclusion}

This paper presented a landmark-only approach to subject-level unknown-identity identification on the ML2HP dataset under a held-out-subject LOSO protocol. By augmenting the original Leap Motion landmark representation with distance- and angle-based geometric descriptors, the proposed pipeline supports both enrolled-subject recognition and rejection of held-out unknown identities under a nested threshold-selection procedure.

The experimental results support three main conclusions. First, the proposed geometric descriptors consistently improve performance over raw landmark coordinates alone, with the full feature set yielding the best results across all evaluated models. Second, among the three compared methods, Extra Trees remains the strongest overall model, outperforming both the embedding MLP and MLP+OpenMax in subject-identification accuracy, rejection quality, and computational efficiency. Third, the score-distribution analysis shows that the main advantage of Extra Trees lies not only in strong closed-set discrimination but also in producing better separated known-versus-unknown rejection scores.

Overall, the results indicate that compact and interpretable landmark-based hand descriptors can provide a strong foundation for contactless hand-based unknown-subject rejection and identification on a small-cohort benchmark. At the same time, the remaining gap between the tree ensemble and the neural baselines suggests that robust rejection of unknown identities remains the central challenge in this setting. The proposed descriptors also reflect a deliberate feature-engineering choice: although the embedding MLP and MLP+OpenMax provide an initial comparison with learned representations, they do not exhaust the range of possible end-to-end or self-supervised representation-learning approaches. Therefore, the superiority of Extra Trees should be interpreted within the present small-cohort, landmark-only LOSO setting rather than as evidence against deep representation learning in general.

Future work will extend the protocol to larger and more diverse cohorts and investigate calibrated metric-learning and end-to-end representation-learning models for improved unknown-subject rejection, including temporal and multi-view architectures that can learn directly from raw landmark sequences.

\bibliographystyle{IEEEtran}
\bibliography{references}

\end{document}